\documentclass[letterpaper]{article}
\usepackage{aaai}
\usepackage{times}
\usepackage{helvet}
\usepackage{courier}
\usepackage{graphicx}
\usepackage{comment}
\usepackage{amsmath,amssymb} 
\usepackage{color}
\usepackage{subfigure}
\usepackage{eso-pic}
\usepackage{algorithm}
\usepackage{algorithmic}
\usepackage{indentfirst}
\usepackage{float}

\begin{document}
%
\title{Region Comparison Network for Interpretable Few-shot Image Classification}
\author{Zhiyu Xue$^1$, Lixin Duan$^1$, Wen Li$^1$, Lin Chen$^2$, Jiebo Luo$^3$\\
$^1$University of Electronic Science and Technology of China\\
$^2$Wyze Labs, Inc. \\
$^3$University of Rochester\\
{\tt\small xzy990228@gmail.com, lxduan@uestc.edu.cn,
liwenbnu@gmail.com,}\\
{\tt\small gggchenlin@gmail.com, jluo@cs.rochester.edu}
}
\maketitle
\begin{abstract}
\begin{quote}
While deep learning has been successfully applied to many real-world computer vision tasks, training robust classifiers usually requires a large amount of well-labeled data. However, the annotation is often expensive and time-consuming. Few-shot image classification has thus been proposed to effectively use only a limited number of labeled examples to train models for new classes. Recent works based on transferable metric learning methods have achieved promising classification performance through learning the similarity between the features of samples from the query and support sets. However, rare of them explicitly consider the model interpretability. For that, in this work, we propose a metric learning based method named Region Comparison Network (RCN), which aims to reveal how few-shot learning works as in a neural network, to learn specific regions that are related to each other in images coming from the query and support sets. Moreover, we design a visualization strategy named Region Activation Mapping (RAM) to intuitively explain what our method has learned by visualizing intermediate variables in our network. We also present a new way to generalize the interpretability from the task level to the category level, which can also be viewed as a way to find the prototypical parts for supporting the final decision of our RCN. Extensive experiments on four benchmark datasets clearly show the effectiveness of our method over existing baselines.
\end{quote}
\end{abstract}

\section{Introduction}
Benefiting from the power of large-scale training data, deep learning models have demonstrated promising performance on many computer vision tasks~\cite{huang2017densely,He2016Deep,szegedy2017inception,Krizhevsky2012Alex,HuSqueeze}. However, it is still a big challenge to apply deep learning to a task with only limited data available, which is often the case in real-world applications. 
As a result, \emph{few-shot learning}, which aims to learn a classifier for a given set of classes with only limited labeled training samples, has been attracting more and more attention from the community in recent years~\cite{huang2019compare,LiMeta,kim2019egnn}.

Many works have been proposed to address the few-shot learning problem based on various principles, \emph{e.g.}, meta learning, metric learning~\cite{snell2017prototypical,VinyalsMatching,sung2018learning}, however, rare attention was paid to the interpretability of few-shot learning models, except for some brand new works~\cite{Explanation-Guided,cao2020concept} in 2020. Although some concrete results are shown in previous works~\cite{satorras2018few,santoro2016meta}, it is still unclear how the model explicitly performs the recognition and comparison process. In other words, we are still confused about the incidence relation between the final classification and the pairs of support and query samples. To this end, in this work, we take one step towards the interpretability of few-shot learning by exploiting the relation between representative regions of different images. We are keen to find answers to the following questions, which parts of a given test(query) image are essential for classification, while which parts of a training(support) sample matter? 

The recognition process of humans partially inspires our method. It is known that human is able to recognize a new object by only seeing a few examples ~\cite{lake2011one,gidaris2018dynamic,qi2018low}. As shown in~\cite{Chen2018ThisLL,sung2018learning}, if we ask humans to describe how they identify objects in the real world, most people might view that focusing on partitions of an image and comparing them with prototypical parts of images from a given category can help them achieve this goal. For example, humans can classify an image of woodpecker mainly because this woodpecker's beak is closely similar to the beaks of woodpeckers they have seen.  

\begin{figure*}[t]
\centering 
\subfigure[]
{
\label{Fig.Concept}
\includegraphics[width=8.7cm,height=5cm]{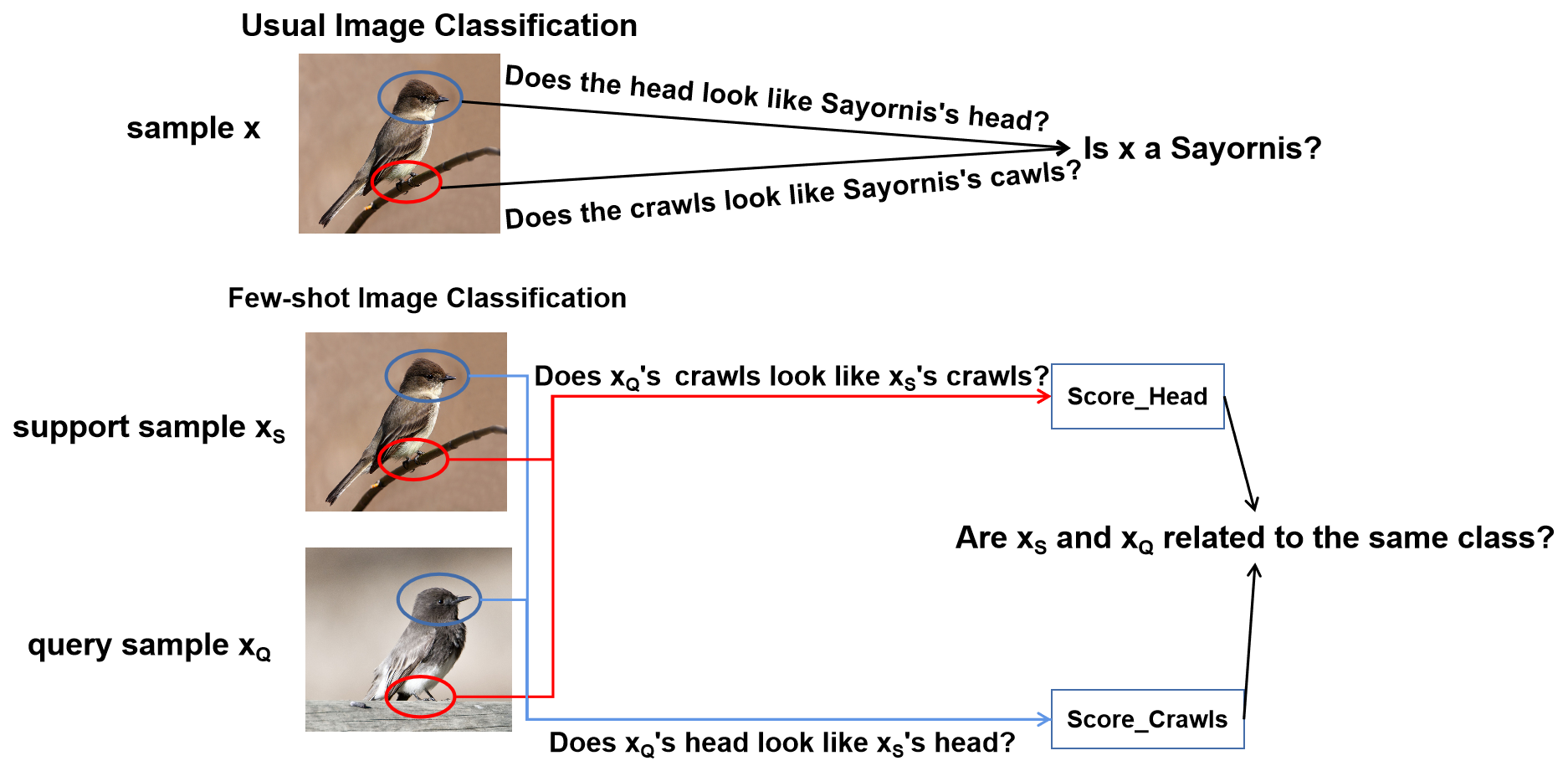}
}
\subfigure[]
{
\label{Fig.Motivation}
\includegraphics[width=8.5cm]{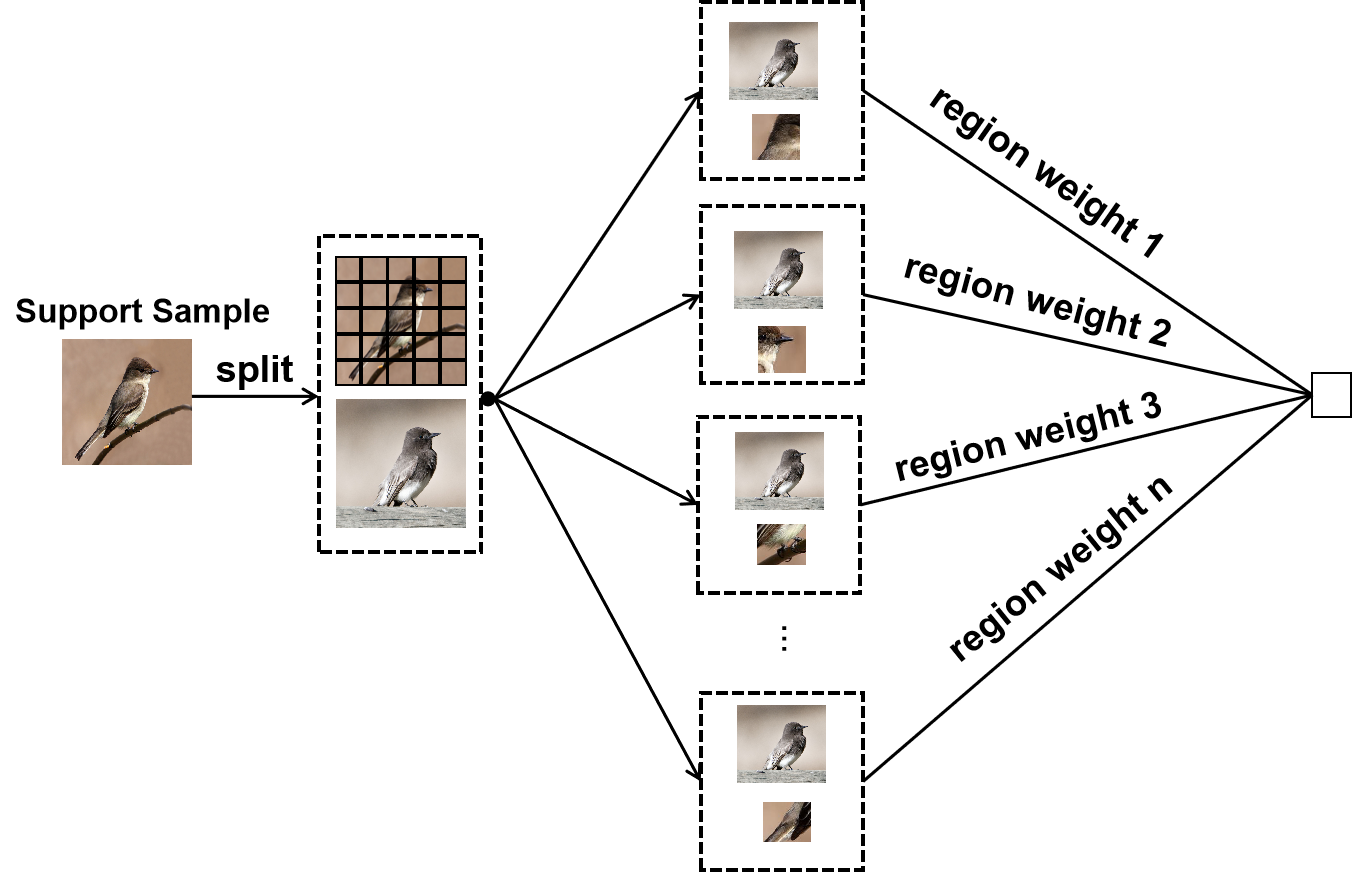}
}
\caption{\ref{Fig.Concept}: 
In traditional image classification tasks, we often explain our reasoning by dissecting the image and pointing out some prototypical parts that can impact the final classification crucially~\cite{Chen2018ThisLL}. However, this theory of reasoning about usual image classification is not suitable for few-shot image classification, since the training strategy and model structure are completely different. We set a new theory for the reasoning in few-shot classification as an ensemble process, which means combining all the region similarity scores into a final similarity score by giving a region weight.\\
\ref{Fig.Motivation}: Our motivation to solve this issue is dividing each support sample into several parts manually. For each query sample, we compute its feature similarity to these parts one by one. In the last procedure, we combine all the region similarity scores into a final classification decision by using a generated weight.}
\end{figure*}

To study this issue, we design a new metric learning based model for few-shot learning. The motivation of our model is to find which parts in a query sample are most similar to the manually selected regions in a support sample, by comparing the computed similarity between them. To achieve this goal, our model is designed to generate a region weight in the final stage, in order to define which common parts between support sample and query sample can influence the final similarity score mostly. Also, we develop Region Activation Mapping (RAM) to acquire some concrete visualization results about interpretability in few-shot image classification, which have rarely been considered in previous works~\cite{satorras2018few,santoro2016meta}. Considering the difference between interpretability in normal image classification and few-shot image classification, it is reasonable to think about what our model can do under the circumstance of data limitation, which means we cannot access sufficient samples to discover the prototypical regions.  Moreover, we also need to find out how much a single region similarity score can contribute to the final similarity score. The difference of interpretability between normal tasks and few-shot tasks can be shown in Fig.~\ref{Fig.Concept}, and our key idea of building the interpretable few-shot learning model can be shown in Fig.~\ref{Fig.Motivation}.

Our contributions can be concluded as follows:
\begin{itemize}
    \item In this paper, we propose a metric learning based model to solve the problem of interpretable few-shot image classification. Compared to attention mechanism, our model indicates the relationship between final classification decision and region similarities directly in the last layer, which can be viewed as a simple and easily explained linear process.  
    \item We present an easily explainable module to make the final prediction for few-shot image classification. By learning a generated weight of regions, this module can explain the question as "what kind of regions in a support sample are similar to somewhere in a query sample, and which of them do the model like to compare?". For that, we develop a so-called region meta learner, which can be viewed as a dynamic system aiming to adapt different meta tasks in the training/testing stage. 
    \item We also present an easy-to-implement visualization strategy named Region Activation Mapping (RAM) to intuitively show the interpretability of our RCN model, by visualizing the weight and similarity scores of regions. We also present a statistic-based method to generalize and quantify the explanations into a set of standard rules for the comparison process, as well as a generalization method to find the prototypes.
\end{itemize}
\begin{figure}
  \includegraphics[width=9cm,height=5cm]{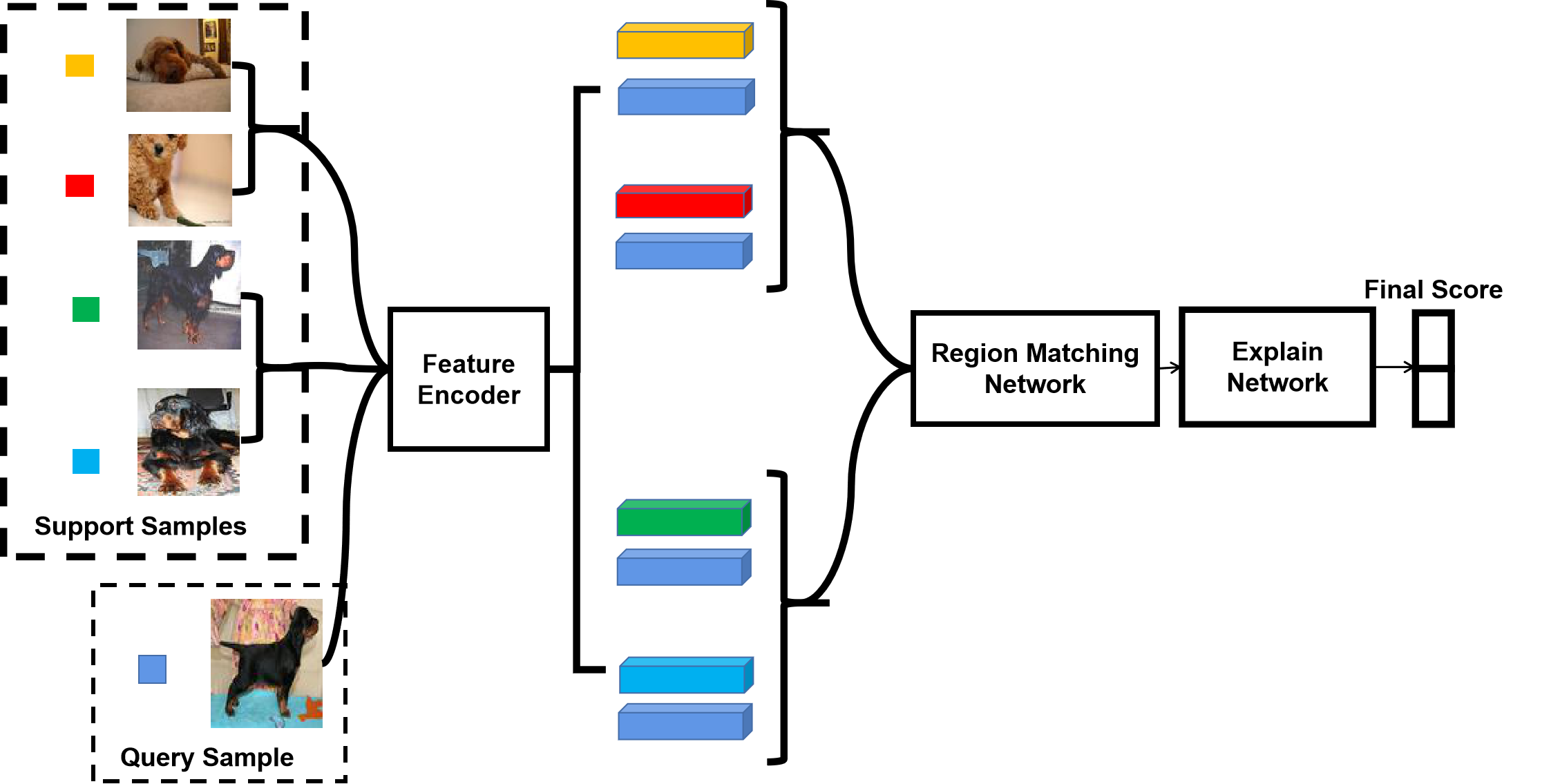}
  \caption{The architecture of 2-way 2-shot}
  \label{fig:Overall}
\end{figure}

\section{Related Works}
\noindent{\bf Few-shot Learning} is a research issue aiming to learn the concept from only few examples per class~\cite{lake2011one}. It requires an efficient representation learning which can extract knowledge from only a few labeled samples and generalize this learned knowledge among many unlabeled samples. It is closely relevant to meta learning~\cite{metagan,hou2019cross}, because we need a model to handle tasks from different tasks. Investigating many recent works of few-shot learning, we group them into metric-based models~\cite{koch2015siamese,VinyalsMatching,snell2017prototypical,sung2018learning} and gradient-based models~\cite{finn2017model,ravi2016optimization,LiMeta}. Metric based methods like matching networks~\cite{VinyalsMatching} address the few-shot classification problem by “learning to compare”~\cite{chen2019closerfewshot}, which means the models can achieve classification score by computing the similarity between support sample and query sample using some metric methods, such as Euclidean distance~\cite{snell2017prototypical}. As for gradient-based methods, like MAML~\cite{finn2017model} and MetaSGD~\cite{LiMeta} aiming to find an appropriate gradient-based optimization method for meta learning, they are usually model agnostic and can be used with some metric learning models to achieve higher performance on few-shot learning tasks.   

Our framework is related to the category of metric-based model. However, not like most exiting methods comparing the features on the level of the whole image, our model tends to compare each region between support sample and query sample, which can explore more fine-grained information and find the critical regions related to the final decision.      

\noindent{\bf Interpretability of Deep Learning} is made to find the crucial factors resulting in the final decision of deep neural networks. Decision models learned on a considerable amount of data produced by humans may lead to unfair and wrong decisions since the training data may contain some human biases and prejudices~\cite{survey_I-ML}. For example, a well-trained cat-dog CNN may classify dog images into the right category successfully. However, the most important foundation may be the same lawn background, not the same dog heads, probably because we collect dog images outdoors while collecting cat images indoor. We need to know what actually happens inside deep neural networks. According to~\cite{rudin2018please}, the current methods of interpretability can be divided into interpretable models~\cite{Chen2018ThisLL,zhang2018interpretable,wang2017bayesian} and model diagnosis~\cite{selvaraju2017grad,simonyan2013gradient}. The objective of model diagnosis is using some visualization methods or sampling functions, such as RISE~\cite{petsiuk2018rise} for visualizing the feature maps and LIME~\cite{ribeiro2016lime} for restructuring a more straightforward model by sampling nearby examples to supersede the original model. On the contrary, some recent works related to interpretable model such as InterpretableCNN~\cite{zhang2018interpretable} and ProtoPNet~\cite{Chen2018ThisLL}, firmly claim it is useless and meaningless to find explanations on black-box models, which is just likely to perpetuate the wrong practice~\cite{rudin2018please}, because standard deep learning models are unexplainable intrinsically no matter what diagnosis methods you use. 

Following the idea of building interpretable models to set a white-box reasoning system for the learning process directly~\cite{rudin2018please}, our model achieves the interpretability by quantifying the contributions of important parts in support sample to the final classification decision.

\section{Methodology}\label{section: Approach}
The training process in few-shot learning aims to learn the concepts from meta training tasks and generalize among meta testing tasks, where the category distributions are entirely disjoint. We can acquire meta tasks by sampling from a big dataset containing various examples such as Mini-ImageNet.

Unlike normal training strategy owning train dataset and test dataset related to a same category distribution. We use episodic training~\cite{VinyalsMatching} paradigm in few-shot learning to minimize the generalization error by sample different meta task per episode. In episodic training, we first split the whole dataset $D$ with $|C|$ classes into meta-training dataset $D_{tr}$ with $|C_{tr}|$ classes and meta-testing dataset $D_{te}$ with $|C_{te}|$ classes, where $|C_{tr}| + |C_{te}| = |C|$ and $C_{tr} \cap C_{te} = \emptyset$. 

For N-way K-shot task in meta-training procedure, we first sample $N$ classes from $C_{tr}$ per episode, and then disjointly sample $K$ examples per class as the support set $S$ and $B$ examples as the query set $Q$, respectively. These two sets can be represented as $D_{S}=\{(x_{i},y_{i})\}_{i=1}^{N\times K}$ and $D_{Q}=\{(x_{i},y_{i})\}_{i=1}^{N\times B}$, where $B$ is a hyperparameter that we need to fix in our experiments. The few-shot learning models can get the basic knowledge on the support set and minimize the empirical error on the query set. 

We use the strategy as same as we mentioned above to evaluate our model on meta-testing dataset $D_{te}$. 

\subsection{Our Approach}\label{subsection: model}
The Region Comparison Network(RCN) is partially inspired by ProtoPNet~\cite{Chen2018ThisLL}. ProtoPNet aims to explain the learning process by comparing the inputting images and some selected prototypical parts of each category. However, instead of projecting the prototypical parts of some class onto the latent training patch by a manual updating rule automatically like ProtoPNet, we use a region meta learner inputted with some representative features for the meta task, to generate a region weight indicating the importance of each region in support sample. This dynamic process can provide different explanations for different meta tasks, which is an ability that ProtoPNet does not have.

The main idea of our model is to compare each selected region in support sample to the whole range of query sample by computing each region similarity score between them, and then find out somewhere in query sample similar to this specific selected region in support sample mostly by using a max pooling kernel. As for interpretability, we consider it as a region weight representing the importance of each corresponding region in support sample compared to query sample in the classification process. In other words, the region weight can help us to point out which similarity between region-to-region can mainly determine the similarity between images-to-images by quantifying the contributions of regions. We achieve this goal by using the region matching network and explaining network that we will introduce in detail in the following section.

Our framework contains three modules: feature extractor, region matching network and explain network. The architecture can be shown in Figure~\ref{fig:Overall}. Feature extractor $f(\cdot)$ is a simple CNN without full-connection layer, which is utilized to map an inputting image into representative feature maps. The region matching network $g(\cdot)$ aims to get the region similarity scores between support sample and query sample, and explain network $h(\cdot)$ can get the final classification decisions by combining the region similarity scores with a weight generated from the region meta learner, which can be taken as an explainable inference process. We will introduce some details of $g(\cdot)$ and $h(\cdot)$ in the following article, respectively. 

For loss function, we use mean square error (MSE) for the loss function of our model(Equal~\ref{E1}). It is not a standard choice for classification problem~\cite{sung2018learning}, but considering our final classification decision is a classification score,it can be taken as a regression problem to achieve our predictions closer to the ground truth generated discretely from $\{0,1\}$. Also, the MSE loss is introduced to measure the gap between the estimated similarity and true similarity of each pair of a query image and a support image, since the similarity is real-valued, we believe the MSE loss is more suitable.

\begin{equation}\scriptsize
    MSE = \sum_{(x_{S},y_{S})\in D_{S}} \sum_{(x_{Q},y_{Q})\in D_{Q}} (s_{S,Q} - 1(y_{S}==y_{Q}))^2\quad \label{E1}
\end{equation}
where $s_{S,Q}$ denotes the final classification score for support sample $x_{S}$ and query sample $x_{Q}$, as well as the similaity between $x_{Q}$ and $x_{S}$.
\subsubsection{Region Matching Network}\label{subsubsection:matching}
\begin{figure}
    \centering
    \includegraphics[width=8.3cm,height=4.7cm]{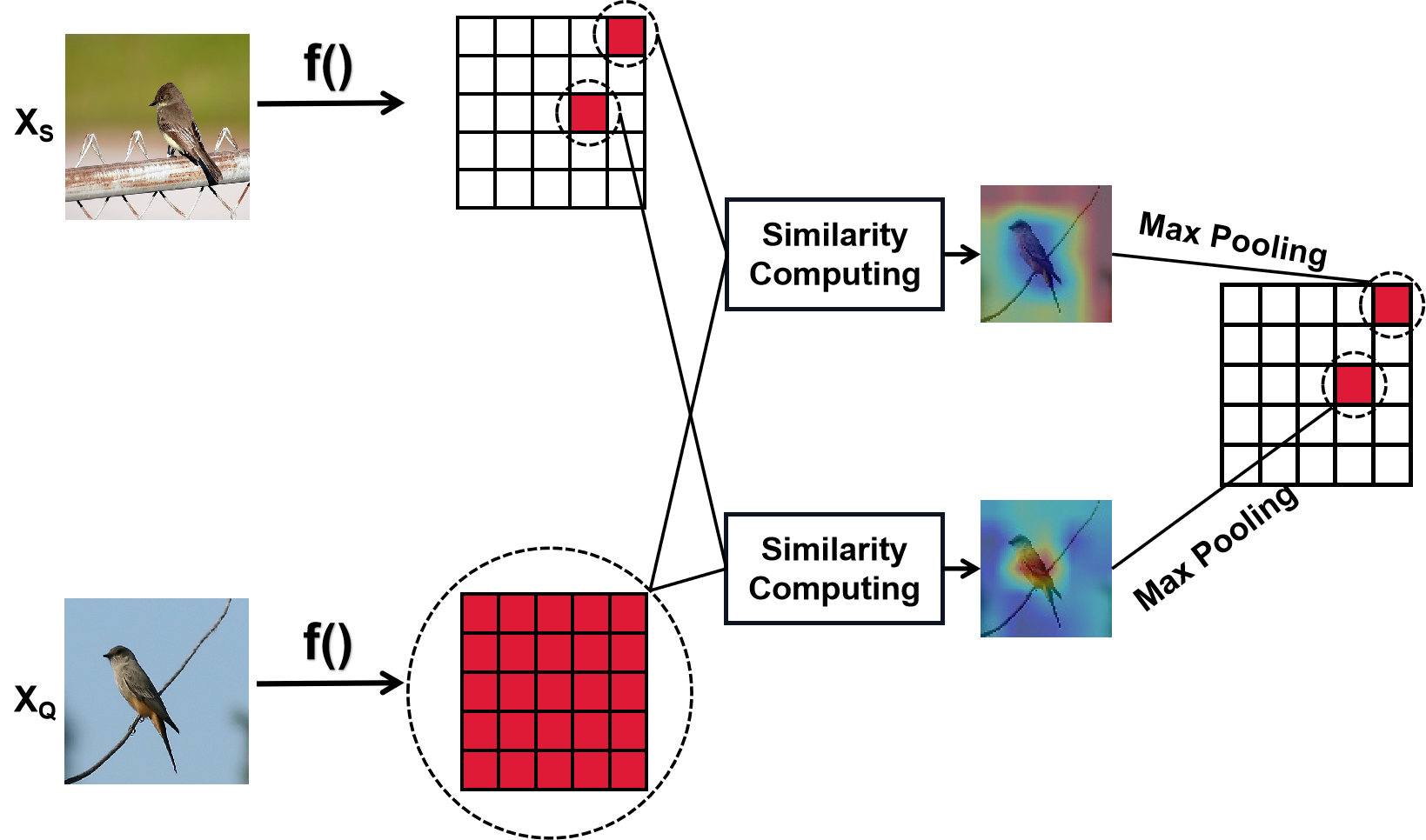}
    \caption{The structure of region matching network for $w=h=5$, where $X_{S}$ and $X_{Q}$ denote support sample and query sample respectively.}
    \label{fig:middle}
\end{figure}

~\\
The Region Matching Network(Figure~\ref{fig:middle}) is built as a method of combination and similarity computing module, which does not have any parameter to learn during the meta training stage. Moreover, the time and space complexity of this module are both lower than that of the regular convolutional layer, which will be explicitly analyzed in our supplementary material. We denote a support sample as $(x_{S},y_{S})$ and a query sample as $(x_{Q},y_{Q})$. The feature maps outputting from the feature extractor $f(\cdot)$ can be represented as $f(x)\in \mathbb{R}^{n\times w \times h}$, where $n,w,h$ represent the number of channels, width and height for feature maps $f(x)$ respectively. 

We first decompose the feature maps $f(x_{S})$ into several region vectors $\{f(x_{S})^{i}\}_{i=1}^{w\times h}$ among width and height, where $f(x_{S})^{i} \in \mathbb{R}^{n\times 1\times 1}$ and $i\in[1,2...w\times h]$. We view it as the representative features of specific regions, which is located in i-th parts of support sample $x_{S}$. For dimensional unification in the similarity computing process, we define a operating function $r(\cdot)$ to repeat a single region vector $f(x_{S})^{i}$ on the dimensionality of width and height, to make them be the same value as those in $f(x_{Q})$. We set this operation as $f'(x_{S})^{i} = r(f(x_S)^{i})$, where $f'(x_{S})^{i} \in \mathbb{R}^{n\times w\times h}$. In order to avoid internal covariate shift, we restrict the similarities into the range of 0 and 1 by utilizing cosine similarity(Eq~\ref{E2}) as the metric method, which measures the similarity of two vector by the cosine of the angle between them. Also, we find it is the best metric function by some empirical study, which will be shown in our supplement material.

\begin{equation}\scriptsize
    CosineSimilarity(a,b) = \dfrac{a\cdot b}{\|a\|\|b\|} \label{E2}
\end{equation}
The region similarity maps $\{S_{S,Q}^{i}\in \mathbb{R}^{1\times w \times h}\}_{i=1}^{h\times w}$ are computed by using cosine similarity between $f'(x_{S})^{i}$ and $f(x_{Q})$on the dimensionality of channels. It can be shown in Eq~\ref{E3} regularly.

\begin{equation}\scriptsize
\begin{aligned}
(S_{S,Q}^{i})_{a,b} = CosineSimilarity((f'(x_{S})^{i})_{a,b},(f(x_{Q}))_{a,b})\\
where\quad f(x)_{a,b}\in \mathbb{R}^{n\times1\times1}, a\in [1,2...w], b\in [1,2...h] \label{E3}
\end{aligned}
\end{equation}

After that, we use a global max pooling kernel to select the most salient information in region similarity maps $\{S_{S,Q}^{i}\}_{i=1}^{h\times w}$, which can be denoted as $\{P_{S,Q}^{i}\}_{i=1}^{h\times w}$. $P_{S,Q}^{i}$ is regarded as a similarity score between $f(x_{S})^{i}$ and somewhere similar to $f(x_{S})^{i}$ mostly in $f(x_{Q})$. Take two bird images for example, $P_{S,Q}^{1}$ may represent how similar the backgrounds are between support sample and query sample, while $P_{S,Q}^{2}$ may represent the similarity of the birds' wings or something else. 

\subsubsection{Explain Network}\label{subsubsection:explain}
Explain Network aims to explain how much that each item in $P_{S,Q}$ contributes to the final classification decisions. In this module, we use a region meta learner to generate the region weight $W_{p}$, and then combine the region similarity scores $P_{S,Q}$ to get the final classification score by using the region weight $W_{p}$.

Considering the important parts are changing from meta-tasks, such as we classify dog images by their heads while birds images by their wings. We utilize a region meta learner to generate a dynamic region weight adapting to each specific meta-task. We will introduce the details of the region meta learner's structure in the experimental section.

Moreover, region meta learner generates region weight by learning from some representative information, which is set as the concatenation of support feature maps and query feature maps on the dimensionality of channel. This process can be represented in Eq~\ref{Eq:explain}.

\begin{figure}
    \centering
    \includegraphics[width=9.1cm,height=4.7cm]{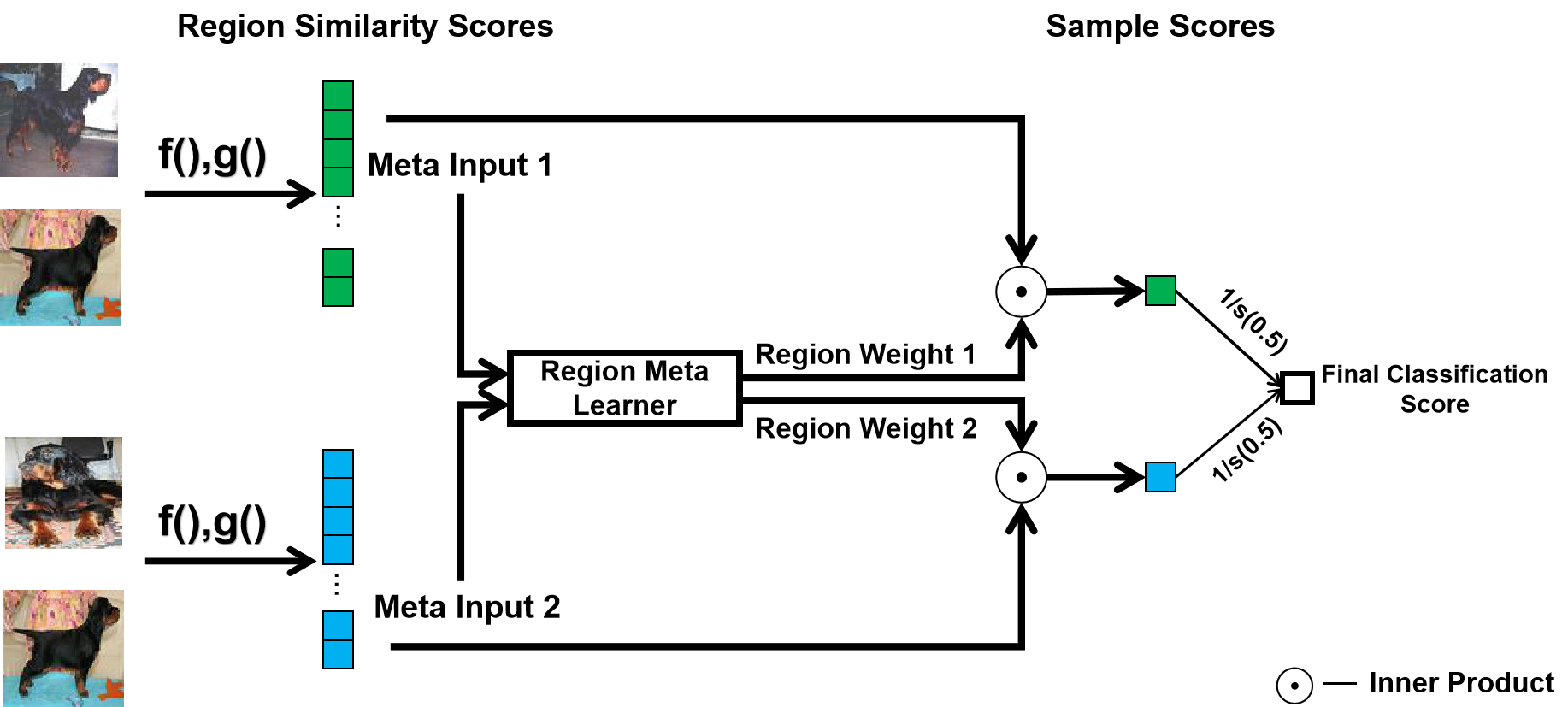}
    \caption{The structure of explain network for 2-shot task(images are from Mini-ImageNet)}
    \label{fig:explain}
\end{figure}

\begin{equation}\scriptsize
\begin{aligned}
&W_{p} = m([f(X_S),f(X_Q)]) \label{Eq:explain}\\
&h(P_{S,Q}) = W^{\mathsf{T}}_{p}P_{S,Q}
\end{aligned}
\end{equation}
where $h(\cdot)$ denotes the explain network and $m(\cdot)$ denotes the region meta learner that we will introduce its structure carefully in the Section~\ref{section: Experiments}. 

{\bf Why do we not use a learnable linear liner layer to acquire region weight, but use a meta learner to generate the region weight instead?} It is mainly because the meta task in each episode is different. For example, we may identify sparrows by their heads, but we identify woodpeckers mainly by their beaks. A simple linear hidden layer may not be able to generalize among different support-query pairs(meta tasks), while using a mete learner instead can alleviate this problem, since it can generate different region weights by giving different meta inputs adapting to different meta tasks. We will demonstrate this assumption by the experimental ablation results in Section~\ref{subsec:ablation}. 

\begin{table*}\small
\begin{center}
\caption{Mean accuracies (\%) of different methods on the MiniImageNet and CIFAR-FS dataset. Results are obtained over 600 test episodes with 95\% confidence intervals. Note that Conv4-n denotes 4-layer convolution network outputting feature maps with n channels. *: \cite{wu2019parn} uses feature extractor as 6-layer convolution network with deformable convolution kernel~\cite{dai2017deformable}}
\scalebox{0.9}{
\begin{tabular}{c c c c c c c}
  \hline
   Model&Backbone&Type&\multicolumn{2}{c}{Mini-ImageNet (5-way)}&\multicolumn{2}{c}{CIFAR-FS (5-way)}\\
   & & & 1-shot & 5-shot & 1-shot & 5-shot\\\hline
  META LSTM~\cite{ravi2016optimization} & Conv4-32 & Meta &  43.44$\pm$0.77 &  60.60$\pm$0.71 & - & - \\
  MAML~\cite{finn2017model}& Conv4-32 & Meta & 48.70$\pm$1.84 &  63.11$\pm$0.92 & 58.9$\pm$1.9 & 71.5$\pm$1.0\\
  Dynamic-Net~\cite{gidaris2018dynamic} & Conv4-64 & Meta & 56.20$\pm$0.86 & 72.81$\pm$0.62 & - & - \\
  Dynamic-Net~\cite{gidaris2018dynamic} & Res12 & Meta & 55.45$\pm$0.89 & 70.13$\pm$0.68 & - & - \\
  SNAIL~\cite{mishra2017simple} & Res12 & Meta & 55.71$\pm$0.99 & 68.88$\pm$0.92 & & \\
  AdaResNet~\cite{lee2019meta}& Res12 & Meta & 56.88$\pm$0.62 & 71.94$\pm$0.57 & - & - \\
  \hline
  MATCHING NETS~\cite{VinyalsMatching} & Conv4-64 & Metric & 43.56$\pm$0.84 &  55.31$\pm$0.73 & - & - \\
  PROTOTYPICAL NETS~\cite{snell2017prototypical} & Conv4-64 & Metric & 49.42$\pm$0.78 &  68.20$\pm$0.66 & 55.5$\pm$0.7 & 72.0$\pm$0.6\\
  RELATION NETS~\cite{sung2018learning} & Conv4-64 & Metric & 50.44$\pm$0.82 &  65.32$\pm$0.70 & 55.0$\pm$1.0 & 69.3$\pm$0.8  \\
  GNN~\cite{satorras2018few} & Conv4-64 & Metric &50.33$\pm$0.36 &  66.41$\pm$0.63 & 61.9 & 75.3 \\
  PABN~\cite{huang2019compare} & Conv4-64 & Metric & 51.87$\pm$0.45&  65.37$\pm$0.68 & - & - \\
  TPN~\cite{liu2018learning}& Conv4-64 & Metric &  52.78$\pm$0.27 &   66.59$\pm$0.28 & - & - \\
  DN4~\cite{li2019revisiting} & Conv4-64 & Metric & 51.24$\pm$0.74 & 71.02$\pm$0.64 & - & - \\
   R2-D2~\cite{bertinetto2018meta} & Conv4-512 & Metric & 51.80$\pm$0.20  & 68.4$\pm$0.20 & 65.3$\pm$0.2 & 79.4$\pm$0.1 \\
   GCR~\cite{li2019gcr} & Conv4-512 & Metric & 53.21$\pm$0.40 & 72.32$\pm$0.32 & - & - \\
  
   PARN~\cite{wu2019parn} & * & Metric & 55.22$\pm$0.82 & 71.55$\pm$0.66 & - & -\\
  \hline
   RCN& Conv4-64 & Metric & 53.47$\pm$0.84 & 71.63$\pm$0.70 & 61.61$\pm$0.96 & 77.63$\pm$0.75\\
   RCN& Res12& Metric & \textbf{57.40$\pm$0.86} & \textbf{75.19$\pm$0.64} & \textbf{69.02$\pm$0.92} & \textbf{82.96$\pm$0.67} \\
  \hline
\end{tabular}}
\label{table: mini}
\end{center}
\end{table*}

\section{Experiments}\label{section: Experiments}

\subsection{Datasets}
To compare our proposed framework with exiting state-of-art few-shot learning methods, We evaluate our proposed framework on four benchmark datasets. The four datasets are introduced as follows: 

\noindent{\bf Mini-Imagenet}~\cite{VinyalsMatching}
 is a dataset containing 60,000 colorful images coming from 100 classes, with 600 images in each class, and it can be taken as a subset of ImageNet~\cite{deng2009imagenet}. In our experiments, we use the same splits of~\cite{snell2017prototypical}, who employ 64 classes for meta-training, 16 for meta-validation and 20 for meta-testing.

\noindent{\bf CIFAR-FS}~\cite{bertinetto2018meta} is randomly sampled from CIFAR-100~\cite{krizhevsky2009learning} by applying the same criteria in~\cite{bertinetto2018meta} as same as MiniImageNet, which means we split the 100 classes to 64 classes for meta-training, 16 for meta-validation and 20 for meta-testing. 

\noindent{\bf CUB-200}~\cite{welinder2010cub} is a fine-grained with 6033 images from
200 bird species. Due to the different split method, we perform experiments following~\cite{li2019revisiting} (130 classes for meta-training, 20 classes for meta-validation and 50 classes for meta-testing) and~\cite{chen2019closerfewshot} (100 classes for meta-training, 50 classes for meta-validation and 50 classes for meta-testing), respectively.

\noindent{\bf Stanford Dogs}~\cite{stanforddogs} contains 20580 images with 120 classes of dogs. Without loss of generality, we use the same criterion in~\cite{li2019revisiting} to split it to few-shot dataset, which means 70,20 and 30 classes for meta-training, meta-validation and meta-testing, respectively.

\subsection{Implementation Details}
\noindent{\bf Feature Extractor}:
We use ResNet-12 following~\cite{lee2019meta,mishra2017simple} and Conv4( a standard 4-layer convolutional network with 64 filters per layer)~\cite{sung2018learning,li2019revisiting,VinyalsMatching} as our feature extractor. Both of them have been used extensively.For ResNet-12, we use DropBlock regularization~\cite{ghiasi2018dropblock} with $keep\_rate = 0.5$ to prevent overfitting. 

\noindent{\bf Region Matching Network}:
In region matching network, the values of width and height of the inputting feature maps $f(x)$ are 5. In the ablation experiments, we use an adaptive average pooling kernel to change the size of feature maps to 1 and 3, in order to find the best outputting size for $f(x)$. 

\noindent{\bf Explain Network}:
For region meta learner $m(\cdot)$, we use a simple CNN to generate the region weights from the concatenation of query feature maps and support feature maps. We take the block of this CNN as $[Conv(1\times1,in\_channels,out\_channels),BN,Relu]$, where $in\_channels$ and $out\_channels$ respectively denote the numbers of channels for input feature maps and output feature maps, and $BN$ is batch normalization. We stack these blocks as the channels of $640\to64$ and $64\to1$. 

\noindent{\bf Data Argumentation}:
Data argumentation is an effective trick to prevent overfitting in training deep learning models. In our experiments, we only apply data argumentation methods on the query set in the meta-training stage. We use a group of random resize crop, random color jittering, random horizontal flip and random erasing~\cite{zhong2020RandomE} as our data argumentation method.

\noindent{\bf Optimization}: 
Adam is used as the optimization method in the meta-training stage. The learning rate is initially set to 0.001 and later reduces to 0.5 times if the average accuracy on the meta-validation dataset over 600 episodes does not increase. The model is trained following a strategy that set an iteration as 500 meta-training episodes, 600 meta-validation episodes and 600 meta-testing episodes. 

\subsection{Results}

\begin{table}[t]\small
\begin{center}

\caption{Mean accuracies (\%) of different methods on the CUB-200. Results are obtained over 600 test episodes with 95\% confidence intervals. $\dag$: Split CUB as~\cite{li2019revisiting}. $\ddag$: Split CUB as~\cite{chen2019closerfewshot}}
\scalebox{0.56}{
\begin{tabular}{c c c c c}
  \hline
  Model&Backbone&Type&\multicolumn{2}{c}{CUB-200 (5-way)}\\
   & & & 1-shot & 5-shot\\\hline
    PCM$^\dag$~\cite{wei2019pcm} & Conv4-64 & Metric & 42.10$\pm$1.96 &  62.48$\pm$1.21 \\
    MATCHING NETS$^\dag$~~\cite{VinyalsMatching} & Conv4-64 & Metric & 45.30$\pm$1.03 &  59.50$\pm$1.01 \\
    PROTOTYPICAL NETS$^\dag$~~\cite{snell2017prototypical} & Conv4-64 & Metric &  37.36$\pm$1.00 & 45.28$\pm$1.03 \\
    GNN$^\dag$~~\cite{satorras2018few} & Conv4-64 &Metric & 51.83$\pm$0.98 &  63.69$\pm$0.94\\
    DN4$^\dag$~~\cite{li2019revisiting} & Conv4-64 &Metric & 53.15$\pm$0.84 & 81.90$\pm$0.60\\
    \hline
    RCN$^\dag$ & Conv4-64 &Metric & 66.48$\pm$0.90 & 82.04$\pm$0.58 \\
    RCN$^\dag$ & Res12 &Metric & \textbf{78.64$\pm$0.88} & \textbf{90.10$\pm$0.50} \\
    \hline
    \hline
    Baseline++$^\ddag$~\cite{chen2019closerfewshot} & Res10 & Metric & 
    69.55$\pm$0.89 & 85.17$\pm$0.50 \\
    MAML++(High-End)+SCA$^\ddag$~\cite{maml++sca} & - & Meta & 70.46$\pm$1.18 & 85.63$\pm$0.66 \\
    GPShot(CosSim)$^\ddag$~\cite{patacchiola2019GPShot} & Res10 & Meta & 70.81$\pm$0.52 & 83.26$\pm$0.50 \\
    GPShot(BNCosSim)$^\ddag$~\cite{patacchiola2019GPShot} & Res10 & Meta & 72.27$\pm$0.30 & 85.64$\pm$0.29  \\

  \hline
  RCN$^\ddag$& Conv4-64 &Metric & 67.06$\pm$0.93 & 82.36$\pm$0.61\\
   RCN$^\ddag$& Res12 & Metric & \textbf{74.65$\pm$0.86} & \textbf{88.81$\pm$0.57} \\
  \hline
\end{tabular}
\label{table:CUB}
}
\end{center}
\end{table}
We sample 15 query images per class for evaluation in both 1-shot and 5-
shot tasks following~\cite{sung2018learning}, and the final few-shot classification accuracies
are computed by averaging over 600 episodes in meta-testing stage. Some meta learning models need to pretrain on a lager task of N-way K-shot before training on 5-way 5-shot(1-shot), which is called meta pretraining~\cite{snell2017prototypical}. Moreover, some models use self-supervised pretraining~\cite{mangla2019charting} or pertrained feature extractor~\cite{lee2019meta}. However, our framework can be meta-trained end-to-end without any method of pretraining.

We  present  the  results  of  our  method  comparing other baselines on the normal datasets and  fine-grained datasets(Tables~\ref{table: mini} \ref{table:CUB} and \ref{table: Dog})

\begin{table}[t]
\begin{center}
\caption{Mean accuracies (\%) of different methods on the Stanford Dogs. Results are obtained over 600 test episodes with 95\% confidence intervals.
} 
\scalebox{0.53}{
\begin{tabular}{c c c c c}
  \hline
  Model&Backbone&Type&\multicolumn{2}{c}{CUB-200 (5-way)}\\
   & & & 1-shot & 5-shot\\\hline
    PCM~\cite{wei2019pcm} & Conv4-64 & Metric & 28.78$\pm$2.33 &  46.92$\pm$2.00 \\
    MATCHING NETS~\cite{VinyalsMatching} & Conv4-64 & Metric & 45.30$\pm$1.03 &  59.50$\pm$1.01 \\
    PROTOTYPICAL NETS~\cite{snell2017prototypical} & Conv4-64 & Metric & 37.59$\pm$1.00 & 48.19$\pm$1.03\\
    GNN~\cite{satorras2018few} & Conv4-64 &Metric & 46.98$\pm$0.98 & 62.27$\pm$0.95\\
    DN4~\cite{li2019revisiting} & Conv4-64 &Metric & 45.73$\pm$0.76 & 66.33$\pm$0.66 \\
    \hline
    RCN & Conv4-64 &Metric & 54.29$\pm$0.96 & 72.65$\pm$0.72\\
    RCN & Res12 &Metric & \textbf{66.24$\pm$0.96} & \textbf{81.50$\pm$0.58} \\
    \hline
\end{tabular}
\label{table: Dog}
}
\end{center}
\end{table}

We find our framework can both achieve promising performances on normal datasets and fine-grained datasets, especially for fine-grained datasets. Due to the motivation of comparing the specific regions instead of the whole images, our model can explore more fine-grained information of each sample and superior the state-of-the-art baselines on the task of few-shot fine-grained image classification. Also, our model is very easy-implemented, since the structures of the region matching network and explain network are simple and can be trained end-to-end on only one training stage.

\subsection{Ablation Study}\label{subsec:ablation}
In order to demonstrate it is correct and reasonable for using a meta learner to generate region weight and also in order to find the impact of the feature maps' size(width and height) to the final classification, we did controlled experiments including using fixed linear layer, learnable linear layer and meta learner. For fixed linear layer, we just add all the region similarity scores by mean operation. For learnable linear layer, all the values of weight items need to be bigger than 0 during the optimization stage. We use an average pool layer to control the height and width of the feature maps.

We use Mini-ImageNet and CUB-200 representing normal benchmarks and fine-grained benchmarks, respectively. The results can be shown in Table~\ref{table: ablation}. We will show the results of ablation experiments of other two datasets(CIFAR-FS and Stanford Dogs) in the supplementary materials. 

\begin{table}[h]\tiny
\begin{center}
\label{table: ablation}
\caption{Mean accuracies (\%) of different methods on the Mini-ImageNet and CUB-200(using split criterion as~\cite{li2019revisiting}). Results are obtained over 600 test episodes with 95\% confidence intervals. Note that the items in the region weight of fixed linear layer are all fixed as $\frac{1}{h\times w}$} \label{table: ablation}
\begin{tabular}{c c c c c}
  \hline
   Version&\multicolumn{2}{c}{Mini-ImageNet}&\multicolumn{2}{c}{CUB-200}\\
   &1-shot & 5-shot & 1-shot & 5-shot \\ \hline
  \hline
  Fixed Linear Layer (5$\times$5) & 49.30$\pm$0.89  & 55.51$\pm$0.71  & 62.61$\pm$1.63 & 67.26$\pm$0.83  \\
  Learnable Linear Layer (5$\times$5) & 55.97$\pm$0.86  & 72.80$\pm$0.63  & 73.23$\pm$0.90 & 88.12$\pm$0.56  \\
  Meta Learner (5$\times$5) & \textbf{57.40$\pm$0.86} & \textbf{75.19$\pm$0.64} & \textbf{78.64$\pm$0.88} & \textbf{90.10$\pm$0.50}\\
  \hline
  Fixed Linear Layer (4$\times$4)& 51.79$\pm$0.90 & 57.40$\pm$0.70 & 65.18$\pm$1.08 & 71.65$\pm$0.83\\
  Learnable Linear Layer (4$\times$4) & 55.18$\pm$0.84 & \textbf{73.25$\pm$0.64} & 75.12$\pm$0.89 & 87.63$\pm$0.54  \\
  Meta Learner (4$\times$4)& \textbf{55.73$\pm$0.83} & 72.78$\pm$0.62 & \textbf{76.48$\pm$0.86} & \textbf{87.89$\pm$0.57} \\
  \hline
  Fixed Linear Layer (3$\times$3)& 51.51$\pm$0.90 & 56.02$\pm$0.70 & 65.97$\pm$1.03 & 74.59$\pm$0.89 \\
  Learnable Linear Layer (3$\times$3) & \textbf{56.50$\pm$0.87}  & \textbf{73.48$\pm$0.62}  & \textbf{76.15$\pm$0.87} & \textbf{88.10$\pm$0.51}  \\
  Meta Learner (3$\times$3)& 55.41$\pm$0.85 & 72.16$\pm$0.68 & 75.63$\pm$0.88 & 86.96$\pm$0.57 \\
\hline
  Fixed Linear Layer (2$\times$2)& 51.58$\pm$0.91 & 57.59$\pm$0.70 & 68.95$\pm$1.05 & 77.64$\pm$0.81 \\
  Learnable Linear Layer (2$\times$2) & \textbf{56.03$\pm$0.85} & 72.23$\pm$0.64 & 73.79$\pm$0.85 & \textbf{87.42$\pm$0.57}  \\
  Meta Learner (2$\times$2)& 55.65$\pm$0.83 & \textbf{72.36$\pm$0.64} & \textbf{75.79$\pm$0.87} &  86.64$\pm$0.55\\
  \hline
  Fixed Linear Layer (1$\times$1)& 52.22$\pm$1.03 & 57.34$\pm$0.75 & 70.70$\pm$0.78 & 78.43$\pm$0.43 \\
  Learnable Linear Layer (1$\times$1) & 54.80$\pm$0.86  & 71.80$\pm$0.69  & \textbf{75.83$\pm$0.85} &  \textbf{86.97$\pm$0.53}  \\
  Meta Learner (1$\times$1)& \textbf{55.40$\pm$0.89} &\textbf{72.78$\pm$0.62}  & 73.83$\pm$0.98 & 84.77$\pm$0.54 \\
  \hline
\end{tabular}
\end{center}
\end{table}

According to the table~\ref{table: ablation}, we can prove that using a region meta learner to generate different region weights for different support-query pairs can improve our model's performance in some cases like $5\times5$, and we will demonstrate that it can improve the interpretability by showing some visual results in Section~\ref{section: Vis}. However, we find the meta learner cannot outperform than learnable linear layer in some cases where the width and height are smaller than 5. It can be reasoned as the adaptive average pool layer may cause some loss of representative information. 

\begin{figure}
\centering 
\subfigure[]
{
\label{Fig.Vis_1}
\includegraphics[width=8.0cm]{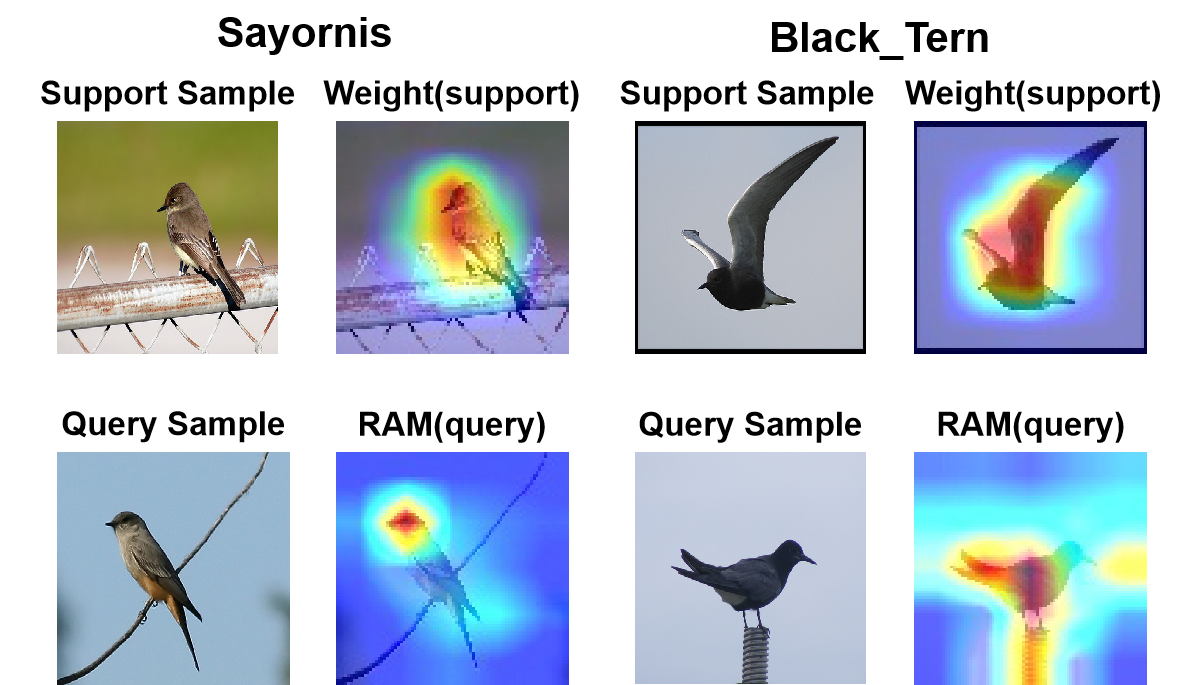}
}
\subfigure[]
{
\label{Fig.Vis_2}
\includegraphics[width=8.0cm]{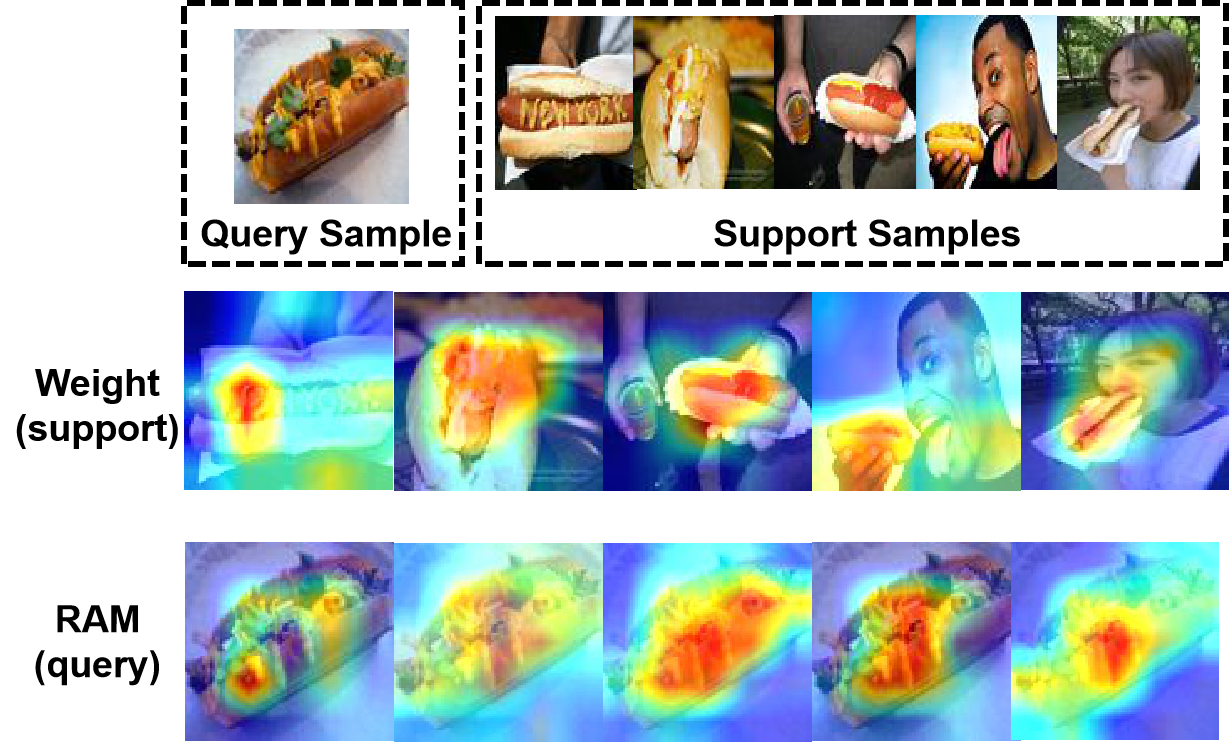}
}
\caption{\ref{Fig.Vis_1}: 
We show examples of class Sayornis and class Black Tern in CUB-200. According to the visualization results about RAM for query sample and region weight for support sample, it is safe to claim that our model classifies these two Saynornis images (left) mainly depending on their heads. As for Black Tern images (right), we find out that our model mainly makes the final classification decision by comparing their wings. This figure shows that our model tends to focus on different important regions for different kinds of images.\\        
\ref{Fig.Vis_2}: 
We show the samples in category number n07697537 of Mini-ImageNet, which can be considered as the category of hot dog bun. Unlike CUB-200 or Stanford Dogs, whose samples almost locate the main objects in the center of the images with plain backgrounds, the main objects in the images of Mini-ImageNet are more difficult to be located and recognized(Like a man holds a hot dog bud located in the left corner). Our model can still find the essential regions due to the help of region meta learner. Moreover, we can explain the issue that our model classifies the hot dog bud images by comparing the sausages.}
\label{fig:visualization}
\end{figure}

As for the size of feature maps and the number of regions we select from the support sample, it needs to be trade-off. In other words, this hyper-parameter must be a value that is not too big or too small. If the size is too small, we cannot capture the important regions and explore the fine-grained information, while we may have too much noise region vectors like the regions of backgrounds if the size is too big.  

\section{Visualization of Model Interpretability}\label{section: Vis}
Attribution methods~\cite{petsiuk2018rise,zhou2016learning} focus on explaining neural networks by finding which parts of the input samples are the most responsible for determining the output of model. When they are applied to deep convolutional models, they can output saliency maps pointing out the important regions in input image~\cite{fong2019pertubation}. 
To make our explanation more comprehensive and user-friendly, we present an easy-implemented visualization method named Region Activation Mapping (RAM) to show the important regions in query sample. Also, the important regions of support sample can be shown by region weight $W_{p}$.

In Class Activation Mapping (CAM)~\cite{zhou2016learning}, the authors hold the view that the feature maps located in different channels focus on different regions in input feature maps, and they use a weight to average them into a saliency map for characterizing the import regions. In RAM, it is reasonable to say that the $i$-th region similarity map $S^{i}_{S,Q}$ represents the similarities between the i-th region in support sample and everywhere in query sample. Therefore, we can make ensemble all the region similarity maps by region weight to indicate the important areas in query sample.

In RAM, we denote the region weight generated by region meta learner as $W_{p}\in \mathbb{R}^{h\times w}$ and the region similarity maps $\{s_{i}\}_{i=0}^{h\times w}$ where $s_{i}\in R^{1\times h\times w}$. Our method can be shown in Eq~\ref{E:4}, where $W_{p}[i]$ denotes the $i$-th item in $W_{p}$. It is easy to find that our final classification decisions for query samples are influenced by the region weight and region similarity maps together so that RAM can show this combined influence very clearly. Note that $k(\cdot)$ is a nonlinear function to enhance the impact of similar regions between query sample and support sample. Here it is set as $k(x) = e^{2x}$.      
\begin{equation}\scriptsize
    RAM = \sum_{i=1}^{h\times w}W_{p}[i]\cdot k(S_{S,Q}^{i})\label{E:4}
\end{equation}
We use RAM to visualize the important regions in query samples, while use region weight for the visualization of support sample. In addition, similarity maps are applied to find out the regions in query sample similar to the determinate regions in support sample. These results are shown in Fig.~\ref{fig:visualization}. In this figure, bilinear upsampling is used to match the input image's size and the results in the visualization process. 

The Figure~\ref{Fig.Vis_1} can demonstrate that our model can focus on different important regions for different categories, while the Figure~\ref{Fig.Vis_2} can prove that our model can locate different regions for different images from the same category. Due to the limitation of pages, we will show more visualization results including the similarity maps in our supplement materials.

\section{Generalization and Quantification of Model Interpretability}\label{section: Generalization}
Our framework can provide the interpretability for a specific meta task, which means it can only explain which regions are essential in a single episode. Therefore, we cannot find which parts are important among the class level, as well as a prototype part or a common rule for classification. 

In this section, we apply an algorithm based on some statistical analysis to generalize the interpretability from meta tasks into categories, and we also present a criterion to metric the importance of regions at the level of class.

We assume a sample set $\{(x_{i},y_{i})\}_{i=1}^{N}$  related to a specific category $c$($y_{i}=c$). We select one sample randomly as a support sample $x_{S}$, while other images are considered as query samples $\{x^{i}_{Q}\}^{N-1}_{i=1}$. We compute the region weight $W_{i}$ between $x_{S}$ and $x^{i}_{Q}$ iteratively, and stack them into a region weight matrix $W \in \mathbb{R}^{(N-1)\times w\times h}$, where $W_{i,j}$ denotes the $j$-th item in $i$-th region vector $W_{i}$. 
If $W_{:,j}$ is a zero vector, we will remove this vector from matrix $W$, since it denotes the $j$-th region in support sample is meaningless. 

We compute the mean value $\mu_{j}$ and the standard deviation $\sigma_{j}$ of vector $W_{:,j}$, and the distribution of $W_{:,j}$ is assumed as a Gaussian distribution. 
We use the probability density function of one-dimensional Gaussian distribution to simulate the distribution of $W_{:,j}$, which can be represented as Eq~\ref{E:Gaussian}:
\begin{equation}\scriptsize
    f(x,W_{:,j}) = \dfrac{1}{\sigma_{j}\sqrt{2\pi}}\exp{(-\frac{(x-\mu_{j})^2}{2\sigma_{j}^2})} \label{E:Gaussian}
\end{equation}
where $\mu_{j}$ and $\sigma_{j}$ denote the mathematical expectation and the standard deviation, respectively. 

It is safe to say that $\mu_{j}$ represents the importance of $j$-th region among the whole class, while $\sigma_{j}$ denotes the degree of dispersion for each support-query pair. Therefore, if $\mu_{j}$ is bigger and $\sigma_{j}$ is smaller, the $j$-th selected region in $x_{S}$ will be more likely to be a prototypical part representing the whole class $c$. We present a criterion about the importance of a region by using the mathematical expectation of $N(\mu_{j},\sigma_{j})$ in range of $[\mu_{j}-2a,\mu_{j}+2a]$, 
This indicator can describe how much the $j$-th region in the selected support image can represent the decision basis of the whole class $c$. It is shown in Eq~\ref{E:Indicator} more minutely:
\begin{equation}\scriptsize
\begin{aligned}
     I_{j} &= \int_{\mu_{j}-2a}^{\mu_{j}+2a} f(x,W_{:,j})x \mathrm{d}x\label{E:Indicator}\\
    a &= \frac{1}{M}\sum_{j}\sigma_{j}
\end{aligned}
\end{equation}
where $a$ is a mean value for standard deviations, which can be taken as a standard deviation for all the similarity values of the selected regions in support sample. 

Our generalization method can be shown in Alg~\ref{alg:Gen} briefly, where $f(\cdot)$, $g(\cdot)$ and $m(\cdot)$ denote the feature extractor, region matching network and region meta learner respectively. In this algorithm, we finally rank $\{I_{j}\}_{j=1}^{M}$ in ascending order.

\begin{algorithm}
\caption{Generalization Method}
\label{alg:Gen}
\begin{algorithmic}[1]
\REQUIRE $x_{S}$, $\{x^{i}_{Q}\}^{N-1}_{i=1}$
\ENSURE $\{I_{j}\}_{j=1}^{M}$
\STATE $W = []$ is a two-dimensional matrix
\STATE $M = 0$
\FOR{$x_{Q}^{i} \in \{x^{i}_{Q}\}^{N-1}_{i=0}$} 
\STATE $S_{i} = m(g(f(x_{S}),f(x_{Q}^{i})))$
\IF{$S_{i} \ne \vec{0}$}
\STATE $W = [W;S_{i}]$
\STATE $M+=1$
\ELSE 
\STATE continue
\ENDIF
\ENDFOR 
\FOR{$j \in [1,2,...M]$}
\STATE $I_{j}=\int_{\mu_{j}-2a}^{\mu_{j}+2a} f(x,W_{:,j})x \mathrm{d}x$
\ENDFOR
\end{algorithmic}
\end{algorithm}

In order to show the results, we take a class in CUB-200 as an example(Fig.~\ref{Fig:Analysis}), and we will show the examples of other datasets(e.g. MiniImageNet) in the supplementary material. 

\begin{figure}
    \centering
    \includegraphics[width=8.5cm,height=6.5cm]{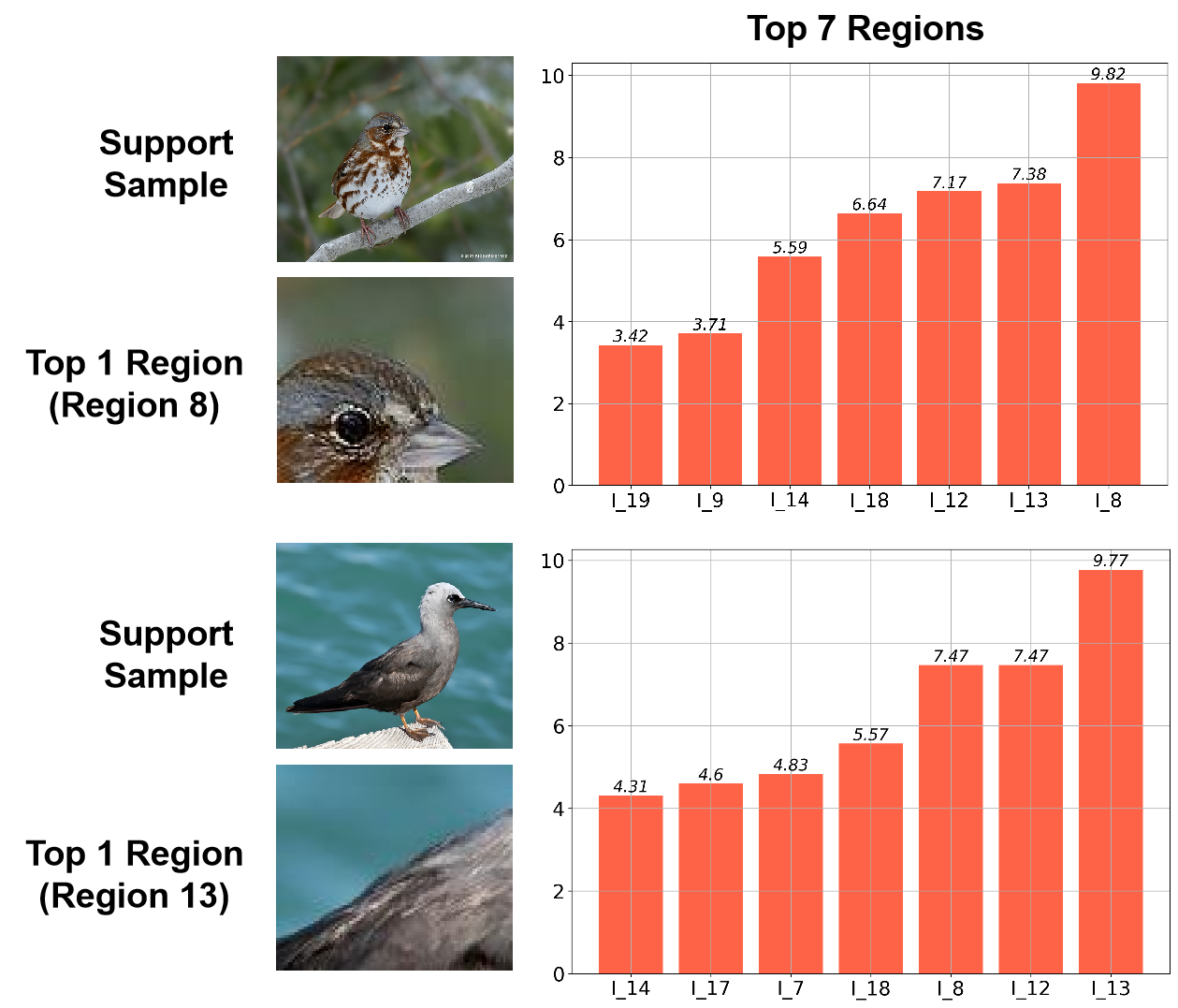}
    \caption{This figure uses the examples from class fox sparrow and class black tern in CUB-200 with the size of $w=h=5$. We select two support samples from these two different categories, and apply our generalization method on them. The maximal $I_{j}$ can be shown clearly in the bar graph(region 8 and region 7). Obviously, they are different parts of the objects, where head part is the prototype~\cite{Chen2018ThisLL} for class fox sparrow and tail part for black tern. In other words, our model tends to classify different objects by different parts, just as same as humans.}
    \label{Fig:Analysis}
\end{figure}

According to Fig.~\ref{Fig:Analysis}, the results of interpretability among the whole class are reasonable and do not against our common sense. Through this generalization method, we can explain which parts of images that the model would like to pay attention to at the level of category, as well as general types of rules to classify images. In addition, it is also a diagnostic method to determine whether our model has focused on the reasonable and explainable areas that do not against our common sense. 

\section{Conclusion}
In this paper, we present an interpretable deep learning framework named Region Comparison Network (RCN) to solve the problem of few-shot image classification. We also present a simple yet useful visualization method named Region Active Mapping (RAM) to show the intermediate variables of our network, which intuitively explains what RCN has learned. Moreover, we present a criterion to measure the importance of regions in each category and develop a strategy to generalize the quantitative explanations from a specific support-query pair to the whole class. Experiments on four benchmark datasets demonstrate the effectiveness of our RCN. Since little work on the explicit interpretability of few-shot learning has been focused on in the literature, we believe our pioneer work is important and can pave the way for future study on this topic.

\bibliographystyle{aaai}
\bibliography{sample-base}

\newpage
\section{Supplementary Material}
\subsection{Different Metrics}
The results of empirical study about using different metric methods in Region Matching Network. 

\begin{table}[H]
    \caption{Mean accuray(\%) of different metric methods using in region matching network, which are evaluated on MiniImageNet. $d(a,b)$: the euclidean distance between vector $a$ and $b$}
    \label{table: different metrics}
    \centering
    \begin{tabular}{c c c}
         \hline
         Metric Methods&\multicolumn{2}{c}{Mini-ImageNet(5-way)}  \\
         &1-shot&5-shot \\ \hline
         Cosine Similarity&57.40$\pm$0.86&\textbf{75.19$\pm$0.64} \\
         \hline
         Tanimito Index&\textbf{57.63$\pm$0.87}&74.31$\pm$0.67 \\
         \hline
         $e^{-d(a,b)}$&56.77$\pm$0.87&72.54$\pm$0.66 \\
         \hline
         $\frac{1}{1+d(a,b)}$&54.61$\pm$0.86&70.27$\pm$0.63 \\
         \hline
    \end{tabular}
 
\end{table}

where tanimito index is represented as $T(a,b)=\dfrac{a\cdot b}{||a||\cdot ||b||-a\cdot b}$.

\subsection{Complexity Analysis of Region Matching Network}
The time and space complexity of Region Matching Network(RMN) is low compared to other layers in the backbone networks. Let us assume the size of the feature map from the backbone networks is $C*H*W$, where $C$, $H$ and $W$ denote the number of channels, height, and width, respectively. Then, the time complexity is $O(H*W*C*W*H)$ and space complexity is $O(H*W*W*H)$ for storing the similarity matrix.

To compare with a normal convolutional layer, assuming its output size equals to the input size, the time complexity is $O(H*W*K*K*C1*C2)$, where $K$, $C1$, $C2$ denotes the kernel size, the numbers of in and out channels, respectively. Note that the size of the feature map is usually small (e.g., $H=W=5$, for Res12), and so $O(H*W*C*W*H)$ is often not as large as $O(H*W*K*K*C1*C2)$. The space complexity of RMN $O(W*H*H*W)$ is also not high, as the space complexity of an ordinary feature map is $O(W*H*C)$ where $C$ can be quite large at high layers (e.g., $C=640$, for Res12).

\end{document}